\documentclass[10pt,twocolumn,letterpaper]{article}

\usepackage{wacv}
\usepackage{times}
\usepackage{epsfig}
\usepackage{graphicx}
\usepackage{amsmath}
\usepackage{amssymb}
\usepackage{booktabs}
\usepackage[table,xcdraw]{xcolor}
\usepackage{import}
\usepackage{mathrsfs}
\usepackage{adjustbox}
\usepackage{multirow}
\usepackage{amssymb}
\usepackage{pifont}% http://ctan.org/pkg/pifont
\newcommand{\cmark}{\ding{51}}%
\newcommand{\xmark}{\ding{55}}%
\usepackage{color}

\def\wacvPaperID{200} % Enter the WACV Paper ID here

\wacvalgorithmstrack   % Uncomment this line if you are submitting to the Algorithms Track.
\wacvfinalcopy % *** Uncomment this line for the final submission
\usepackage[accsupp]{axessibility} % Improves PDF readability for those with disabilities.

\ifwacvfinal
\usepackage[breaklinks=true,bookmarks=false]{hyperref}
\else
\usepackage[pagebackref=true,breaklinks=true,colorlinks,bookmarks=false]{hyperref}
\fi

\begin{document}

%%%%%%%%% TITLE
\title{Dissecting Deep Metric Learning Losses for Image-Text Retrieval}

\author{Hong Xuan, 
Xi (Stephen) Chen\\
Microsoft\\
{\tt\small \{Hong.Xuan|Chen.Stephen\}@microsoft.com}
% \and
% Xi Chen\\
% Microsoft \\
% {\tt\small chnxi@microsoft.com}
}

\maketitle
\thispagestyle{empty}

%%%%%%%%% ABSTRACT
\begin{abstract}
   Visual-Semantic Embedding (VSE) is a prevalent approach in image-text retrieval by learning a joint embedding space between the image and language modalities where semantic similarities would be preserved. The triplet loss with hard-negative mining has become the de-facto objective for most VSE methods. Inspired by recent progress in deep metric learning (DML) in the image domain which gives rise to new loss functions that outperform triplet loss, in this paper we revisit the problem of finding better objectives for VSE in image-text matching. Despite some attempts in designing losses based on gradient movement, most DML losses are defined empirically in the embedding space. Instead of directly applying these loss functions which may lead to sub-optimal gradient updates in model parameters, in this paper we present a novel Gradient-based Objective AnaLysis framework, or \textit{GOAL}, to systematically analyze the combinations and reweighting of the gradients in existing DML functions. With the help of this analysis framework, we further propose a new family of objectives in the gradient space exploring different gradient combinations. In the event that the gradients are not integrable to a valid loss function, we implement our proposed objectives such that they would directly operate in the gradient space instead of on the losses in the embedding space. Comprehensive experiments have demonstrated that our novel objectives have consistently improved performance over baselines across different visual/text features and model frameworks. We also showed the generalizability of the GOAL framework by extending it to other models using triplet family losses including vision-language model with heavy cross-modal interactions and have achieved state-of-the-art results on the image-text retrieval tasks on COCO and Flick30K. Code is available at: \textcolor{magenta}{https://github.com/littleredxh/VSE-Gradient.git}
\end{abstract}

%%%%%%%%% BODY TEXT
\begin{figure}[t]
    \centering
    \includegraphics[width=0.95\columnwidth]{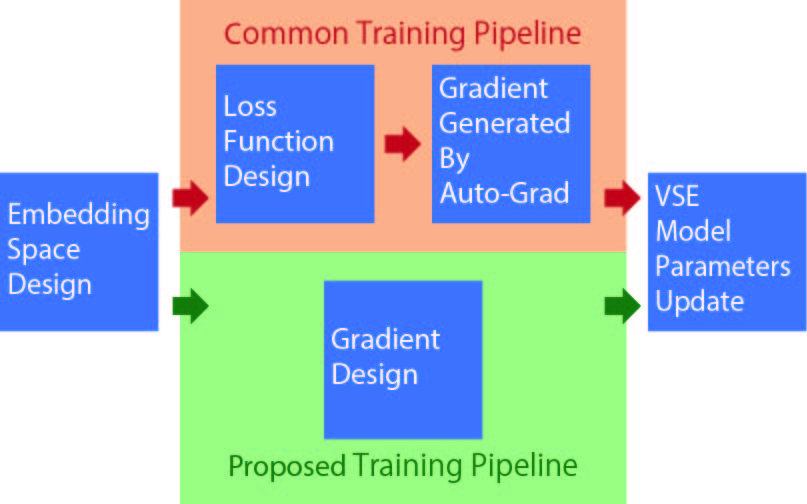}
    \caption{To realize a desired visual semantic embedding space, a common method is to design a loss function which can be calculated on deep learning platforms such as PyTorch or TensorFlow. The auto-grad mechanism on these platforms automatically calculates the gradients to update the model parameters to form a desired embedding space. In practice, the goal of visual semantic embedding is about optimizing the clustering or separation of feature points extracted from image and text, and the loss function is a somewhat indirect approach to reach that goal, while the gradient more directly affects the update of the embedding space. We propose a method to directly design the gradient to train models.
    }
    \label{fig:main}
\end{figure}
\section{Introduction}
Recognizing and describing the visual world with language is a basic human ability but still remains challenging for artificial intelligence. With recent advances in Deep Neural Networks, tremendous progress has been made in bridging the vision-language modalities. Visual-semantic embedding (VSE)~\cite{frome2013devise,kiros2014unifying,faghri2018vse++} is one of the major topics to build a connection between images and natural language. It aims to map images and their descriptive text information into a joint space, such that a relevant pair of image and text should be mapped close to each other while an irrelevant pair of image and text should be mapped far from each other. In this paper, we focus on visual-semantic embedding for the task of image-text matching and retrieval, but our approach is generalizable to other image-text retrieval models using the triplet loss family~\cite{scan,chen2020uniter,li2020oscar,xvlm}.

A VSE model usually consists of feature extractors for image and text, a feature aggregator~\cite{chen2021vseinfty}, and an objective function during training. Despite significant advances of VSE in feature extractors~\cite{vaswani2017attention,dosovitskiy2020vit,butd_anderson2018bottom} and feature aggregators~\cite{wang2020consensus,chen2021vseinfty}, there is less attention on the loss function for training the model. A hinge-based triplet ranking loss with hard-negative sampling~\cite{facenet,faghri2018vse++} has become the de-facto training objective for many VSE approaches~\cite{scan,li2020oscar,zhang2021vinvl}. Few innovations have been made in designing the loss function for learning joint image-text embeddings since then. 

On the other hand, designing deep metric learning (DML) losses has been well-studied for image-to-image retrieval. Many loss functions have been proposed to improve the training performance on image embedding tasks, showing that triplet loss is not optimal for general metric learning~\cite{yi2014deep,SOP,wang2019multi,Xuan_2020_ECCV,Sun_2020_CVPR}. Early losses such as triplet loss and contrastive loss~\cite{facenet,Npairs} are defined with the intuition that positive pairs should be close while negative pairs should be apart in the embedding space. However, such defined loss functions may not lead to desirable gradients which can explicitly impact the update of model parameters. Some attempts have been made in defining the loss function to achieve desirable gradient updates~\cite{yi2014deep,Sun_2020_CVPR}. However, such approaches lack a systematic view and analysis of the combinations in gradients, and are only limited to integrable gradients so that the resulting losses are differentiable. Therefore, these loss functions may not be optimal and applicable to the image-text retrieval task. 

Instead of directly applying established loss functions to VSE for image-text matching, in this paper we present Gradient-based Objective AnaLysis framework, or \textbf{GOAL}, a novel gradient-based analysis framework for the VSE problem. We firstly propose a new gradient framework to dissect the losses at the gradient level, and extract their key gradient elements. Then, we explore a new training idea to directly define the gradient to update the model in each training step instead of defining the loss functions, as shown in Figure~\ref{fig:main}. This new framework allows us to simply combine the key gradient elements in DML losses to form a family of new gradients and avoids the concern of integrating the gradient into a loss function. Finally, the new gradients continue to improve existing VSE performance on image-text retrieval tasks.

In brief, our contributions can be summarized as the following:
\begin{itemize}
    \item We propose a general framework \textbf{GOAL} to comprehensively analyze the update of gradients of existing deep metric learning loss functions and apply this framework to help find better objectives for the VSE problem.
    \item We propose a new method to deal with image-text retrieval task directly by optimizing the model with a family of gradient objectives instead of using a loss function.
    \item We show consistent improvement over existing methods, achieving state-of-the-art results in image-text retrieval tasks on COCO datasets.
\end{itemize}

%%%%%%%%%%%%%%%%%%%%%%%%%%%%%%%%%%%%%%%%%%%%%%%
%% Related works by Stephen/Hong
\section{Related Work}
% 	• VSE method and its variance
% 	• Transformer method for VL
% 	• DML
% 	• Gradient modification
% Related papers ``Do Lessons from Metric Learning.."
%     This work only discusses the three limit loss functions in DML, vinillar triplet loss, NT-Xent loss and SmoothAP loss. 
    
%     The major lesson is that the more negative examples contributed to gradient during optimization, the lower evaluation score. And finally, Triplet loss with Semi-hard negative mining always performance best in the experiments.
    
%     similar observation is also report in EPHN paper.

\textbf{Visual Semantic Embedding for Image-text matching} There is a rich line of literature focused on mapping visual and text modalities to a joint semantic embedding space for image-text matching~\cite{frome2013devise,kiros2014unifying,faghri2018vse++,scan,wu2019unified,chen2021vseinfty}. VSE++ is proposed in~\cite{faghri2018vse++} as a fundamental VSE schema where visual and text embeddings are pretrained separately then aggregated with AvgPool after being projected to a shared space, which later are jointly optimized by a triplet loss with hard-negative mining. Since then consistent advances have been made to improve visual and text feature extractors~\cite{resnet,dosovitskiy2020vit,hochreiter1997lstm,vaswani2017attention,devlin-etal-2019-bert} and feature aggregators~\cite{huang2018learning,li2019visual,wang2020consensus,wu2019unified}. In contrast to dominant use of spatial grids of the feature map as visual features, bottom-up attention~\cite{butd_anderson2018bottom} has been introduced to learn visual semantic embeddings for image-text matching, which is commonly realized by stacking the region representations from pretrained object detectors~\cite{scan,zhang2021vinvl}. ~\cite{chen2021vseinfty} proposed Generalized Pooling Operators (GPO) to learn the best pooling strategy which outperforms approaches with complex feature aggregators. Inspired by the success of large-scale pretraining in language models~\cite{devlin-etal-2019-bert,liu2019roberta}, there is a recent trend of performing task-agnostic vision-language pretraining (VLP) on massive image-text pairs for generic representations, then fine-tune on task-specific data and losses to achieve state-of-the-art results in downstream tasks including image-text retrieval~\cite{lu2019vilbert,tan2019lxmert,chen2020uniter,li2020oscar,xvlm}. However, as opposed to our proposed method, prevalent approaches choose to optimize the triplet loss as the de-facto objective for the image-text matching task. In this paper, we will strive to revisit the problem of finding better training objectives for visual semantic embeddings.

\textbf{Deep Metric Learning} is useful in extreme classification settings such as fine-grained recognition~\cite{SOP,ICR,CUB200,CAR196,facenet}. The goal is to train networks to map semantically related images to nearby locations and unrelated images to distant locations in an embedding space. There are many loss functions that have been proposed to solve the deep metric learning problem. Triplet loss function~\cite{hoffer2015deep,facenet} and its variants such as circle loss~\cite{Sun_2020_CVPR} form a triplet that contains anchor, positive and negative instances, where the anchor and positive instance share the same label, and anchor and negative instance share different labels. Pair-wise loss functions such as contrastive loss~\cite{hadsell2006dimensionality}, binomial deviance loss~\cite{yi2014deep}, lifted structure loss~\cite{SOP} and multi-similarity loss~\cite{wang2019multi} penalize when the distance is large between a pair of instances with the same labels and when the distance is small between a pair of instances with different labels. All these loss functions encourage the distance of positive images pairs to be smaller than the distance of the negative images pairs. Due to the fact that the training goal of DML is similar to VSE problem, in this paper, we borrow these loss design ideas of DML to improve the VSE problem.

\textbf{Gradient Modification} Recent works in DML such as Multi-Similarity Loss and Circle Loss~\cite{wang2019multi,Sun_2020_CVPR,Xuan_2020_ECCV} start with standard triplet loss formulations and adjust the gradients of loss functions to give clear improvements with very simple code modifications.  These works all find explicit loss functions whose gradients are desirable. Other strategies start with a desired gradient weighting function and integrate the desired gradients to derive a loss function that comes with gradients of appropriate properties.  This is often limited to simple weighting strategies, such as the simple linear form in~\cite{Sun_2020_CVPR} and simple gradient removal for positive pairs when triplets contain hard negative in~\cite{Xuan_2020_ECCV}, because it may be hard to find the loss function whose gradient is consistent with complex weighting strategies. The most related work is P2Sgrad~\cite{zhang2019p2sgrad}, which analyzes the gradient in the family of margin-based softmax loss and directly modifies the gradient with the cosine similarity for better optimization. Comparing to P2Sgrad, our work focuses on the triplet loss and its variant loss functions.

The framework in this paper directly explores the space of desired gradient updates. By not limiting ourselves to designing a loss function with appropriate gradients, we can be more explicit in experimentally dissecting the effects of different parts of the gradient.  Furthermore, we can recombine the gradient terms that are experimentally most useful in a form of gradient surgery~\cite{yu2020gradient} that very slightly alters existing algorithms to give improved performance.
%%%%%%%%%%%%%%%%%%%%%%%%%%%%%%%%%%%%%%%%%%%%%%%

\section{Gradient-based Objective Design Framework}
We define a collection of terms for how a batch of images and texts affect a network. Let 
$\mathbf{X}$ be a batch of input images, 
$\mathbf{Y}$ be a batch of input texts, 
$\mathbf{x}$ be the $L2$ normalized feature vectors of the images extracted with the image extractor, 
$\mathbf{y}$ be the $L2$ normalized feature vectors of the texts extracted with the text extractor,  
$l$ be the loss value for the batch, 
$\theta$ be the parameters of the image extractor,
$\phi$ be the parameters of the text extractor,
$\eta$ be the learning rate,
$f_{\theta}(\cdot)$ be the mapping function of the image extractor,
$g_{\phi}(\cdot)$ be the mapping function of the text extractor,
and $L(\cdot)$ be loss function. In the forward training step, the expression is:
 
\begin{equation}
l=L(\mathbf{x}, \mathbf{y}) \text{,  where  } \mathbf{x}=f_{\theta}(\mathbf{X}) \text{  and  } \mathbf{y}=g_{\phi}(\mathbf{Y})
\end{equation}. 

The image and text extractor weights are updated as:
\begin{equation}
\left\{
\begin{aligned}
\theta^{t+1} = \theta^{t} - \eta\frac{\partial l}{\partial \mathbf{x}}\frac{\partial \mathbf{x}}{\partial \theta}\\
\phi^{t+1} = \phi^{t} - \eta\frac{\partial l}{\partial \mathbf{y}}\frac{\partial \mathbf{y}}{\partial \phi}
\end{aligned} 
\right.
\label{eq:gradient_update}
\end{equation}

These two equations highlight that the updates of extractors parameters are combined with two sets of derivative terms. The first set of derivative terms $\frac{\partial l}{\partial \mathbf{x}}$ and $\frac{\partial l}{\partial \mathbf{y}}$ represent how the change of the image and text embedded features affects the loss, and this is the term explored the most in detail in this work. The second set of derivative terms $\frac{\partial \mathbf{x}}{\partial \theta}$ and $\frac{\partial \mathbf{y}}{\partial \phi}$ represent how the change in the model's parameter affects the embedded features. This term can always be expanded with the multiplication of multiple terms for each layer in a modern deep network with multiple layers because of the derivative chain rule, which is not discussed in the work.  

The first set of derivative terms are always constrained by the analytic form of the loss function. For example, due to the exponential form of lifted structure loss~\cite{SOP} and binomial deviance loss~\cite{yi2014deep}, their derivatives also contain an exponential term. Such a term may cause gradient instability and it is an example of how the design of the loss function can at best, only implicitly control the extractor's learning behavior. %But, if the design goal is to explicitly control the gradient to update model, we have to find derivatives which can be integrated into an analytical loss function. Then the design scope is narrowed down again. 

With the latest deep learning platform such as Pytorch~\cite{pytorch} which supports forward modules with customized gradient backward calculation, instead of depending on the derivative of the loss, we can explicitly define the gradient update based on the proposed GOAL framework to directly impact the extractors learning behavior. In the following discussion, we focus on the particular forms of the first set of terms in many triplet loss functions from DML literature, and then propose to directly define the first set of terms for model training.

\subsection{Gradient Components}
\label{sec:idea}
Given a pair of image and text feature $\mathbf{x}$ and $\mathbf{y}$, when the image feature $\mathbf{x}$ is treated as an anchor, we denote its text hard negative feature $\mathbf{y}'$ mined in the text batches $\mathbf{Y}$; 
when the text feature $\mathbf{y}$ is treated as an anchor, we denote its image hard negative feature $\mathbf{x}'$ mined in the image batches $\mathbf{X}$. Then, we can get two triplets $(\mathbf{x},\mathbf{y},\mathbf{y}')$ and $(\mathbf{y},\mathbf{x},\mathbf{x}')$. In the first triplet, $S_{\mathbf{x},\mathbf{y}}=\mathbf{x}^{T}\mathbf{y}$ and $S_{\mathbf{x},\mathbf{y}'}=\mathbf{x}^{T}\mathbf{y}'$ are the cosine similarity computed as the dot-product of the positive and negative pair of normalized image feature and the normalized text feature. Similar cosine similarity is computed for the second triplet, $S_{\mathbf{y},\mathbf{x}}=\mathbf{y}^{T}\mathbf{x}$ and $S_{\mathbf{y},\mathbf{x}'}=\mathbf{y}^{T}\mathbf{x}'$. Finally, these cosine similarities are input into a symmetric triplet loss function $l=L(S_{\mathbf{x},\mathbf{y}},S_{\mathbf{x},\mathbf{y}'})+L( S_{\mathbf{y},\mathbf{x}},S_{\mathbf{y},\mathbf{x}'})$.

The gradients w.r.t. the image and text feature are:
\begin{equation}
\resizebox{1\hsize}{!}{$
\left\{
\begin{aligned}
\frac{\partial l}{\partial \mathbf{x}} 
&=\frac{\partial l}{\partial S_{\mathbf{x},\mathbf{y}}}
\frac{\partial S_{\mathbf{x},\mathbf{y}}}{\partial \mathbf{x}}+
\frac{\partial l}{\partial S_{\mathbf{x},\mathbf{y}'}}
\frac{\partial S_{\mathbf{x},\mathbf{y}'}}{\partial \mathbf{x}}+
\frac{\partial l}{\partial S_{\mathbf{y},\mathbf{x}}}
\frac{\partial S_{\mathbf{y},\mathbf{x}}}{\partial \mathbf{x}}\\
&=\frac{\partial L(S_{\mathbf{x},\mathbf{y}},S_{\mathbf{x},\mathbf{y}'})}{\partial S_{\mathbf{x},\mathbf{y}}}\mathbf{y}+
\frac{\partial L(S_{\mathbf{x},\mathbf{y}},S_{\mathbf{x},\mathbf{y}'})}{\partial S_{\mathbf{x},\mathbf{y}'}}\mathbf{y}'+
\frac{\partial L(S_{\mathbf{y},\mathbf{x}},S_{\mathbf{y},\mathbf{x}'})}{\partial S_{\mathbf{y},\mathbf{x}}}\mathbf{y}\\
\frac{\partial l}{\partial \mathbf{y}} 
&=\frac{\partial l}{\partial S_{\mathbf{x},\mathbf{y}}}
\frac{\partial S_{\mathbf{x},\mathbf{y}}}{\partial \mathbf{y}}+
\frac{\partial l}{\partial S_{\mathbf{y},\mathbf{x}}}
\frac{\partial S_{\mathbf{y},\mathbf{x}}}{\partial \mathbf{y}}+
\frac{\partial l}{\partial S_{\mathbf{y},\mathbf{x}'}}
\frac{\partial S_{\mathbf{y},\mathbf{x}'}}{\partial \mathbf{y}}\\
&=\frac{\partial L(S_{\mathbf{x},\mathbf{y}},S_{\mathbf{x},\mathbf{y}'})}{\partial S_{\mathbf{x},\mathbf{y}}}\mathbf{x}+
\frac{\partial L(S_{\mathbf{y},\mathbf{x}},S_{\mathbf{y},\mathbf{x}'})}{\partial S_{\mathbf{y},\mathbf{x}}}\mathbf{x}+
\frac{\partial L(S_{\mathbf{y},\mathbf{x}},S_{\mathbf{y},\mathbf{x}'})}{\partial S_{\mathbf{y},\mathbf{x}'}}\mathbf{x}'\\
\end{aligned} 
\right.$}
\label{eq:gradient_general}
\end{equation}

There are two major elements in the above gradients: scalars 
$\frac{\partial L(S_{\mathbf{x},\mathbf{y}},S_{\mathbf{x},\mathbf{y}'})}
{\partial S_{\mathbf{x},\mathbf{y}}}$, 
$\frac{\partial L(S_{\mathbf{x},\mathbf{y}},S_{\mathbf{x},\mathbf{y}'})}
{\partial S_{\mathbf{x},\mathbf{y}'}}$, 
$\frac{\partial L(S_{\mathbf{y},\mathbf{x}},S_{\mathbf{y},\mathbf{x}'})}
{\partial S_{\mathbf{y},\mathbf{x}}}$, 
$\frac{\partial L(S_{\mathbf{y},\mathbf{x}},S_{\mathbf{y},\mathbf{x}'})}
{\partial S_{\mathbf{y},\mathbf{x}'}}$, and the unit gradient directions $\mathbf{x}$, $\mathbf{y}$, $\mathbf{x}'$, $\mathbf{y}'$\footnote{They are all unit vector due to the $L2$ normalization}. 

The difference in triplet loss and its variants is primarily coming from the scalars. In DML literature, there are majorly two sets of scalar forms: the scalars related to both positive and negative pair similarity of a triplet, which we denote as (\textbf{Triplet Weight $T$}), and the scalars related to either positive and negative pair similarity of a triplet, which we denote as (\textbf{Pair Weight $P$}). 
\subsection{Triplet Weights}
\label{sec:w_triplet}
\begin{figure}[t]
    \centering
    \includegraphics[width=.32\columnwidth]{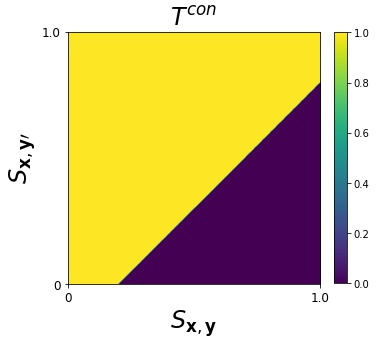}
    \includegraphics[width=.32\columnwidth]{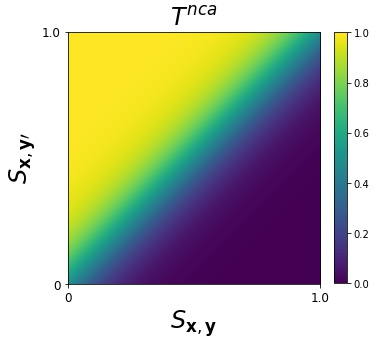}
    \includegraphics[width=.32\columnwidth]{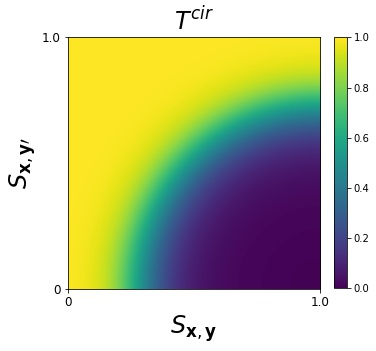}
    \caption{A triplet diagram characterizes the behavior of triplet weights as a function of the the similarity of the positive image-text pair (along the x-axis) and the negative image-text pair (along the y-axis).  Triplets where the anchor, positive and negative features are all very similar will be in the top right of the right, and triplets where the positive pairs are similar and the negative pairs are not similar are in the bottom right corner.  Using this diagram, (left) shows the constant triplet weight $T^{con}$, (middle) shows the NCA triplet weight $T^{nca}$, (right) shows the Circle triplet weight $T^{cir}$.}
    \label{fig:wt}
\end{figure}
For standard triplet loss function with hard negative mining, the gradient can be derived as:
\begin{equation}
\resizebox{1\hsize}{!}{$
\left\{
\begin{aligned}
l^{tri}
&= \max (m+s_{\mathbf{x},\mathbf{y}'}-s_{\mathbf{x},\mathbf{y}}, 0) + \max (m+s_{\mathbf{y},\mathbf{x}'}-s_{\mathbf{y},\mathbf{x}}, 0) \\
\frac{\partial l^{tri}}{\partial \mathbf{x}} 
&= \delta (m+s_{\mathbf{x},\mathbf{y}'}-s_{\mathbf{x},\mathbf{y}}) (\mathbf{y}'-\mathbf{y})-
\delta (m+s_{\mathbf{y},\mathbf{x}'}-s_{\mathbf{y},\mathbf{x}}) \mathbf{y}\\
\frac{\partial l^{tri}}{\partial \mathbf{y}} 
&=-\delta (m+s_{\mathbf{x},\mathbf{y}'}-s_{\mathbf{x},\mathbf{y}}) \mathbf{x}+
\delta (m+s_{\mathbf{y},\mathbf{x}'}-s_{\mathbf{y},\mathbf{x}}) (\mathbf{x}'-\mathbf{x})%\\
% &\frac{\partial l}{\partial \mathbf{x}'} 
% = \delta (m+s_{\mathbf{y},\mathbf{x}'}-s_{\mathbf{y},\mathbf{x}}) \mathbf{y}\\
% &\frac{\partial l}{\partial \mathbf{y}'} 
% = \delta (m+s_{\mathbf{x},\mathbf{y}'}-s_{\mathbf{x},\mathbf{y}}) \mathbf{x}
\end{aligned} 
\right.$}
\label{eq:triplet_loss}
\end{equation}
where $m$ is the margin parameter and $\delta(\cdot)$ is the Heaviside function. 

In the gradients of the triplet loss, all scalars are triplet weights because it contains the similarity of both positive and negative pairs of a triplet. The triplet weight is denoted as constant triplet weight $T^{con}$:
\begin{equation}
T^{con}=\delta (m+s_{\mathbf{x},\mathbf{y}'}-s_{\mathbf{x},\mathbf{y}})
\label{eq:wt_con}
\end{equation}
For simplicity, we only show the weights related to triplet ($\mathbf{x}$, $\mathbf{y}$, $\mathbf{y}'$) in the following discussion, and the discussion of weights for triplet ($\mathbf{y}$, $\mathbf{x}$, $\mathbf{x}'$) is similar. When triplets activate the Heaviside function, $T^{con}$ is a constant $1$, indicating that these eligible triplets will be treated equally. When triplets don't activate the Heaviside function, $T^{con}$ is $0$, indicating that these triplets has no impact on gradient.

A second common loss function is the NT-Xent loss derived from NCA~\cite{nca}, denote as $l^{nca}$. Instead of taking all negative candidates into account, in this paper, we adopt the hard-negative mined version as a fair comparison to triplet loss function. 
\begin{equation} 
\begin{aligned}
l^{nca}=
-[ 
\log(\frac{\exp{(\tau S_{\mathbf{x},\mathbf{y}})}}{\exp{(\tau S_{\mathbf{x},\mathbf{y}})}+\exp{(\tau S_{\mathbf{x},\mathbf{y}'})}})\\
+\log(\frac{\exp{(\tau S_{\mathbf{y},\mathbf{x}})}}{\exp{(\tau S_{\mathbf{y},\mathbf{x}})}+\exp{(\tau S_{\mathbf{y},\mathbf{x}'})}})
]
\end{aligned}
\label{eq:nca_loss}
\end{equation}
where $\tau$ is the scaling parameter. The scalars in its gradient are also a triplet weight which is denoted as NCA triplet weight $T^{nca}$(the derivation is shown in Appendix):
\begin{equation}
T^{nca}=\frac{1}{1+\exp{(\tau(S_{\mathbf{x},\mathbf{y}}-S_{\mathbf{x},\mathbf{y}'}))}}  
\label{eq:wt_nca}
\end{equation}
$T^{nca}$ is rely on the difference of $S_{\mathbf{x},\mathbf{y}}$ and $S_{\mathbf{x},\mathbf{y}'}$. When a triplet in a correct configuration, $S_{\mathbf{x},\mathbf{y}}-S_{\mathbf{x},\mathbf{y}'}>0$, the triplet weight is small. Otherwise, the triplet weight will be large. 

Because $T^{nca}$ only considers the similarity difference $S_{\mathbf{x},\mathbf{y}}-S_{\mathbf{x},\mathbf{y}'}$, some corner cases such as triplet with both large $S_{\mathbf{x},\mathbf{y}}$ and $S_{\mathbf{x},\mathbf{y}'}$ or both small $S_{\mathbf{x},\mathbf{y}}$ and $S_{\mathbf{x},\mathbf{y}'}$ are not well treated. Circle loss~\cite{Sun_2020_CVPR} proposed a circle triplet weight $T^{cir}$ to deal with the cases:
\begin{equation}
T^{cir}=\frac{1}{1+\exp{(\tau(S_{\mathbf{x},\mathbf{y}}(2-S_{\mathbf{x},\mathbf{y}})-S_{\mathbf{x},\mathbf{y}'}^2))}}
\label{eq:wt_cir}
\end{equation}

The idea of $T^{cir}$ is to introduce a non-linear mapping for $S_{\mathbf{x},\mathbf{y}}$ and $S_{\mathbf{x},\mathbf{y}'}$ in the exponential term in order to weight more on the corner cases. 

Figure~\ref{fig:wt} shows the triplet weight diagram, a triplet visualization tool from~\cite{Xuan_2020_ECCV}, for $T^{con}$ with $m=0.2$ and $T^{nca}$ and $T^{cir}$ with $\tau=10$. The equal weight line in $T^{nca}$ is straight lines with form $S_{\mathbf{x},\mathbf{y}}-S_{\mathbf{x},\mathbf{y}'}=\text{const}$. And the equal weight line in $T^{cir}$ is circular lines with form $(S_{\mathbf{x},\mathbf{y}}-1)^2+S_{\mathbf{x},\mathbf{y}'}^2=\text{const}$, demonstrating how it increases the weight to the corner cases. 

\subsection{Pair Weight}
\label{sec:w_pair}
\begin{figure*}[t]
    \centering
    \includegraphics[width=.39\columnwidth]{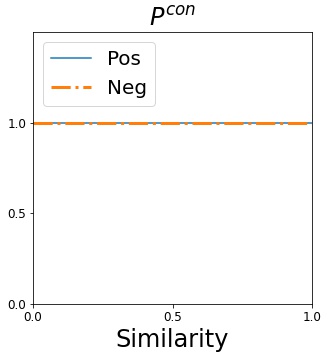}
    \includegraphics[width=.39\columnwidth]{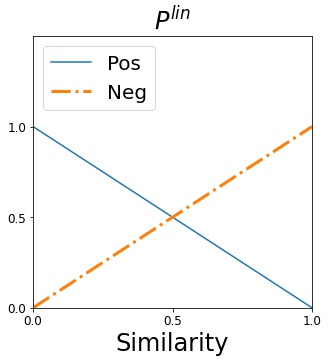}
    \includegraphics[width=.39\columnwidth]{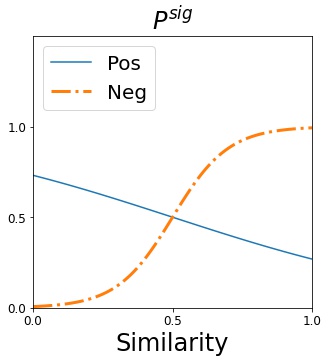}
    \includegraphics[width=.39\columnwidth]{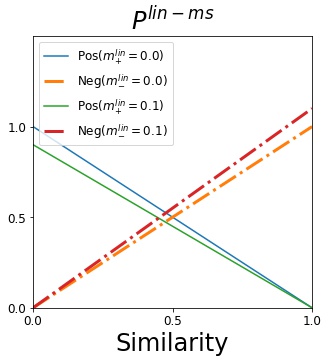}
    \includegraphics[width=.39\columnwidth]{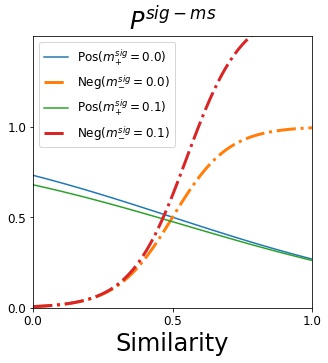}
    \caption{Visualization of constant pair weight $P^{con}$, linear pair weight $P^{lin}$, sigmoid pair weight $P^{sig}$ with $\alpha=2$, $\beta=10$, $\lambda=0.5$ , linear-MS pair weight$P^{lin-ms}$ with $m_{+}^{lin}=m_{-}^{lin}=0.1$ and sigmoid-MS pair weight $P^{sig-ms}$ with $m_{+}^{sig}=m_{-}^{sig}=0.1$}
    \label{fig:wp}
\end{figure*}
In addition to triplet weights, many DML works~\cite{yi2014deep,SOP,wang2019multi,Xuan_2020_ECCV,Sun_2020_CVPR} also proposed pair weights in loss functions. For detailed discussion of pair-weight $P$, we denote the weight of positive pairs $P_{+}$ and the weight of negative pairs $P_{-}$. Let a constant scaling parameter to be a baseline for fair comparison. In this case, both pair weights are set with constant $1$, as:
\begin{equation}
P_{+}^{con}=P_{-}^{con}=1;
\end{equation}

Recent works~\cite{wang2019multi,Xuan_2020_ECCV,Sun_2020_CVPR} argued that the weight for negative pairs should be large when they are close to each other. Otherwise, as mentioned in ~\cite{Xuan_2020_ECCV}, the optimization for DML tasks will quickly converge to a bad local minima. The solution in Circle loss~\cite{Sun_2020_CVPR} is to apply a linear pair weight $P^{lin}$: for negative pairs, the weight is large if the similarity is large and small if the similarity is small; for positive pairs, the weight is large if the similarity is small and small if the similarity is large:
\begin{equation}
\left\{
\begin{aligned}
&P_{+}^{lin}=1-S_{\mathbf{x},\mathbf{y}}\\
&P_{-}^{lin}=S_{\mathbf{x},\mathbf{y}'}
\end{aligned} 
\right.
\label{eq:wp_lin}
\end{equation}

Early work binomial deviance loss~\cite{yi2014deep} uses a similar pair weight but with a nonlinear sigmoid form $P^{sig}$:
\begin{equation}
\left\{
\begin{aligned}
&P_{+}^{sig}=\frac{1}{1+\exp{( \alpha(S_{\mathbf{x},\mathbf{y}}-\lambda))}}\\
&P_{-}^{sig}=\frac{1}{1+\exp{(-\beta (S_{\mathbf{x},\mathbf{y}'}-\lambda))}}
\end{aligned} 
\right.
\label{eq:wp_sig}
\end{equation}
where $\alpha$, $\beta$ and $\lambda$ are three hyper-parameters. 

Multi-similar(MS) loss~\cite{wang2019multi} combines ideas from the lifted structure loss~\cite{SOP} and binomial deviance loss~\cite{yi2014deep}, which includes not only the self-similarity of a selected pair but also the relative similarity from other pairs.  

% The MS paper~\cite{wang2019multi} tries to find a loss function whose derivative fits the proposed pair weight.  Because the relative similarity term involves additional examples (outside the triplet), this creates additional gradients relative to those examples, even though the stated purpose is to weigh the selected pair.  Therefore, it's difficult to understand if the performance gain is coming from the proposed pair weight or from the gradients affecting the feature location of these other examples.  By casting their work within our framework, we can decouple the pair-weighting and explore the impact of this term in isolation.

We follow ~\cite{wang2019multi} to cast their weighting function $P^{sig-ms}$ in our framework.  Given a triplet, the self-similarity of the selected positive pair and negative pair are $S_{\mathbf{x},\mathbf{y}}$ and $S_{\mathbf{x},\mathbf{y}'}$. The similarity of other positives in a batch and negatives to the same anchor is considered as relative-similarity, noted as ${R_{\mathbf{x},\mathbf{y}}}^{i}$ and ${R_{\mathbf{x},\mathbf{y}'}}^{j}$. In addition,~\cite{wang2019multi} also defines $\mathcal{P}$ and $\mathcal{N}$ be the sets of selected ${R_{\mathbf{x},\mathbf{y}}}^{i}$ and ${R_{\mathbf{x},\mathbf{y}'}}^{j}$, where 

\begin{align*}
\mathcal{P}=\{{R_{\mathbf{x},\mathbf{y}}}^{i} \colon {R_{\mathbf{x},\mathbf{y}}}^{i}<\max\{S_{\mathbf{x},\mathbf{y}'}, {R_{\mathbf{x},\mathbf{y}}'}^{j}\}+\epsilon\}\\
\mathcal{N}=\{{R_{\mathbf{x},\mathbf{y}'}}^{j} \colon {R_{\mathbf{x},\mathbf{y}}'}^{j}>\min\{S_{\mathbf{x},\mathbf{y}}, {R_{\mathbf{x},\mathbf{y}}}^{i}\}-\epsilon\} 
\label{eq:set}
\end{align*}

\begin{equation}
\left\{
\begin{aligned}
&P_{+}^{sig-ms}=\frac{1}{m_{+}^{sig}+\exp{( \alpha(S_{\mathbf{x},\mathbf{y}}-\lambda))}}\\
&P_{-}^{sig-ms}=\frac{1}{m_{-}^{sig}+\exp{(-\beta (S_{\mathbf{x},\mathbf{y}'}-\lambda))}}
\end{aligned} 
\right.
\label{eq:wp_sigms}
\end{equation}
where 
\begin{align*}
&m_{+}^{sig}=\frac{1}{\left | \mathcal{P} \right |}\sum_\mathcal{P} \exp{( \alpha (S_{\mathbf{x},\mathbf{y}} - {R_{ap}}^{i}))}\\
&m_{-}^{sig}=\frac{1}{\left | \mathcal{N} \right |}\sum_\mathcal{N} \exp{(-\beta  (S_{\mathbf{x},\mathbf{y}'} - {R_{an}}^{j}))}
\end{align*}

There are two terms in MS loss dynamically changing the pair weight. The self-similarity term has the same effect of sigmoid pair weight $P^{sig}$. As for the relative-similarity term, the major effect is to increase or decrease the maximum magnitude of the pair weight. 

Given a negative pair, when its relative-similarity term $m_{-}^{sig}>1$, this indicates the selected negative example is relatively closer to anchor compared to other negative examples.  Then, the negative weight increases because the relative term decreases the denominator in $P_{-}^{sig-ms}$. When its relative-similarity term $m_{-}^{sig}<1$, indicating the selected negative example is relatively far away from anchor comparing to other negative examples, the negative weight decreases because the relative term increases the denominator in $P_{-}^{sig-ms}$. The latter situation will not exist under the training with hard negative mining. 

Given a positive pair, when its relative-similarity term $m_{+}^{sig}>1$, this indicates the selected positive pair has similarity larger than other positive pairs in its batch, the positive weight decreases because the relative term increases the denominator in $P_{+}^{sig-ms}$. When its relative-similarity term $m_{+}^{sig}<1$, indicating the selected positive pair has similarity less than other positive pairs in its batch, the positive weight increases because the relative term decreases the denominator in $P_{+}^{sig-ms}$. 

When $m_{+}^{sig}=m_{-}^{sig}=1$ the pair weights simplify back to the sigmoid form in equation~\ref{eq:wp_sig}. 

In sum, the main effect caused by the relative-similarity term is to dynamically increase or decrease the maximum penalty for positive and negative pairs as shown in right graph of Figure~\ref{fig:wp}. 

In practice, training MS loss needs to tune four hyper-parameters $\alpha$, $\beta$, $\lambda$ and $\epsilon$ to fit different datasets, making the training not convenient and not efficient. With analysis on relative-similarity terms $m_{+}^{sig}$ and $m_{-}^{sig}$, we define a clearer and parameter free version of pair weight called linear MS pair weight $P^{lin-ms}$, which behaves similar to the original MS weight:
\begin{equation}
\left\{
\begin{aligned}
&P_{+}^{lin-ms}=(1-m_{+}^{lin})(1-S_{\mathbf{x},\mathbf{y}})\\
&P_{-}^{lin-ms}=(1+m_{-}^{lin})S_{\mathbf{x},\mathbf{y}'}
\end{aligned} 
\right.
\label{eq:wp_linms}
\end{equation}where 
\begin{align*}
&m_{+}^{lin}=\frac{1}{\left | \mathcal{P} \right | }\sum_\mathcal{P} (S_{\mathbf{x},\mathbf{y}}-{R_{\mathbf{x},\mathbf{y}}}^{i})\\
&m_{-}^{lin}=\frac{1}{\left | \mathcal{N} \right | }\sum_\mathcal{N} (S_{\mathbf{x},\mathbf{y}'}-{R_{\mathbf{x},\mathbf{y}'}}^{j})
\end{align*}

\subsection{Combinations of Gradient Components}
\begin{table}[t]
\centering
\begin{adjustbox}{width=\columnwidth}
\begin{tabular}{l|c|c|c}
            & $T^{con}$    & $T^{nca}$ & $T^{cir}$ \\
\hline
$P^{con}$   & Triplet loss & NT-Xent loss & New \\
$P^{lin}$   & New & New & Circle loss~\cite{Sun_2020_CVPR} \\
$P^{sig}$   & Binomial deviance~\cite{yi2014deep} & New & New \\
$P^{lin-ms}$& New & New & New \\
$P^{sig-ms}$& MS loss~\cite{wang2019multi} & New & New \\

\hline
\end{tabular}
\end{adjustbox}
\caption{Mapping different gradient combinations of triplet weight and pair weight into existing DML loss functions. Under our GOAL framework, the combinations labeled as ``New" are able to be explored.}
\label{table:methods}
\end{table}
\label{sec:combination}
In this section, we have dissected many previous loss functions from DML in terms of their triplet weights and pair weights. Table~\ref{table:methods} shows how to map different combinations of gradient components into existing loss functions. In addition to these combinations, the remaining combinations labeled as ``New" are all unexplored. These gradient component combinations are hard to be explored if the training needs a loss function and possibly impossible if they are not integrable. However, under our GOAL framework, we are able to train a model with these gradients.

\begin{table}[t]
\centering
\begin{adjustbox}{width=1\columnwidth}
\begin{tabular}{l|c|c}
\hline
& Image$ \rightarrow$ Text 
& Text $\rightarrow$ Image \\
\hline
Method 
& R@1 & R@1 \\
\hline
VSE++(R152,FT) 
& 41.3 & 30.3 \\
VSE++(R152,FT) ours
& 41.0 $\pm$ 0.3 & 30.2 $\pm$ 0.1 \\
\hline
VSE$\infty$(BUTD)
& 58.3 & 42.4 \\
VSE$\infty$(BUTD) ours  
& 58.3 $\pm$ 0.7 & 43.1 $\pm$ 0.0  \\
\hline
VSE$\infty$(WSL)
& 66.4 & 51.6 \\
VSE$\infty$(WSL) ours  
& 66.2 $\pm$ 0.2 & 51.6 $\pm$ 0.3 \\
\hline
\end{tabular}
\end{adjustbox}
\caption{Results verification of the model trained with triplet loss function backward vs the model trained with gradient backward on three VSE methods on COCO dataset. Full table is in Appendix}
\label{table:verify_short}
\end{table}

\begin{table*}[t]
\centering
\begin{adjustbox}{width=1.9\columnwidth}
\begin{tabular}{c|ccc|ccc}
\hline
\multicolumn{7}{c}{VSE++ (ResNet152)}\\
\hline
& \multicolumn{3}{c|}{Image$ \rightarrow$ Text} 
& \multicolumn{3}{c}{Text $\rightarrow$ Image} \\
\hline
& $T^{con}$ & $T^{nca}$ & $T^{cir}$ & $T^{con}$ & $T^{nca}$ & $T^{cir}$ \\
\hline
$P^{con}$    & \cellcolor[HTML]{FFFFFF}33.9 $\pm$ 0.9 & \cellcolor[HTML]{9FD8BC}34.9 $\pm$ 0.4 & \cellcolor[HTML]{E5F5ED}34.2 $\pm$ 0.6 & \cellcolor[HTML]{D2EDE0}22.8 $\pm$ 0.4 & \cellcolor[HTML]{DBF1E6}22.7 $\pm$ 0.2 & \cellcolor[HTML]{FFFFFF}22.3 $\pm$ 0.5 \\
$P^{lin}$    & \cellcolor[HTML]{C8E9D9}34.5 $\pm$ 0.2 & \cellcolor[HTML]{CAEADA}34.5 $\pm$ 0.2 & \cellcolor[HTML]{BFE5D2}34.6 $\pm$ 0.3 & \cellcolor[HTML]{8DD1B0}23.5 $\pm$ 0.1 & \cellcolor[HTML]{A6DBC1}23.2 $\pm$ 0.2 & \cellcolor[HTML]{96D5B6}23.4 $\pm$ 0.4 \\
$P^{sig}$    & \cellcolor[HTML]{9CD7BA}34.9 $\pm$ 0.1 & \cellcolor[HTML]{7BCAA4}35.2 $\pm$ 0.4 & \cellcolor[HTML]{93D4B4}35.0 $\pm$ 0.4 & \cellcolor[HTML]{7ACAA3}23.7 $\pm$ 0.1 & \cellcolor[HTML]{7FCCA6}23.7 $\pm$ 0.2 & \cellcolor[HTML]{84CDA9}23.6 $\pm$ 0.5 \\
$P^{lin-ms}$ & \cellcolor[HTML]{8ED2B1}35.0 $\pm$ 0.5 & \cellcolor[HTML]{77C8A0}35.3 $\pm$ 0.5 & \cellcolor[HTML]{BEE5D2}34.6 $\pm$ 0.4 & \cellcolor[HTML]{77C8A1}23.8 $\pm$ 0.2 & \cellcolor[HTML]{8ED1B0}23.5 $\pm$ 0.2 & \cellcolor[HTML]{A8DCC3}23.2 $\pm$ 0.3 \\
$P^{sig-ms}$ & \cellcolor[HTML]{57BB8A}35.6 $\pm$ 0.1 & \cellcolor[HTML]{9DD8BB}34.9 $\pm$ 0.4 & \cellcolor[HTML]{76C8A0}35.3 $\pm$ 0.4 & \cellcolor[HTML]{57BB8A}24.1 $\pm$ 0.1 & \cellcolor[HTML]{7BCAA4}23.7 $\pm$ 0.2 & \cellcolor[HTML]{80CCA7}23.7 $\pm$ 0.1 \\
\hline
\multicolumn{7}{c}{VSE++ (ResNet152, fine-tuned)}\\
\hline
$P^{con}$    & \cellcolor[HTML]{FFFFFF}40.8 $\pm$ 0.3 & \cellcolor[HTML]{D2EDE0}41.0 $\pm$ 0.3 & \cellcolor[HTML]{ECF8F2}41.2 $\pm$ 0.1 & \cellcolor[HTML]{FDFFFE}30.2 $\pm$ 0.1 & \cellcolor[HTML]{C5E8D7}30.5 $\pm$ 0.0 & \cellcolor[HTML]{FFFFFF}30.1 $\pm$ 0.2 \\
$P^{lin}$    & \cellcolor[HTML]{E6F5EE}41.3 $\pm$ 0.2 & \cellcolor[HTML]{ABDDC5}42.6 $\pm$ 0.3 & \cellcolor[HTML]{B8E3CE}42.3 $\pm$ 0.5 & \cellcolor[HTML]{C7E9D8}30.5 $\pm$ 0.1 & \cellcolor[HTML]{B6E2CC}30.6 $\pm$ 0.1 & \cellcolor[HTML]{A9DCC3}30.7 $\pm$ 0.1 \\
$P^{sig}$    & \cellcolor[HTML]{BCE4D1}42.2 $\pm$ 0.2 & \cellcolor[HTML]{72C69D}43.4 $\pm$ 0.1 & \cellcolor[HTML]{7AC9A3}43.3 $\pm$ 0.0 & \cellcolor[HTML]{6EC59A}31.1 $\pm$ 0.2 & \cellcolor[HTML]{6FC59B}31.1 $\pm$ 0.2 & \cellcolor[HTML]{57BB8A}31.3 $\pm$ 0.2 \\
$P^{lin-ms}$ & \cellcolor[HTML]{D0ECDF}41.8 $\pm$ 0.3 & \cellcolor[HTML]{ABDDC5}42.6 $\pm$ 0.6 & \cellcolor[HTML]{9FD9BD}42.8 $\pm$ 0.2 & \cellcolor[HTML]{B0DFC8}30.7 $\pm$ 0.1 & \cellcolor[HTML]{8BD0AE}30.9 $\pm$ 0.1 & \cellcolor[HTML]{7DCBA5}31.0 $\pm$ 0.2 \\
$P^{sig-ms}$ & \cellcolor[HTML]{64C093}43.6 $\pm$ 0.3 & \cellcolor[HTML]{57BB8A}43.8 $\pm$ 0.5 & \cellcolor[HTML]{59BC8C}43.8 $\pm$ 0.5 & \cellcolor[HTML]{A1D9BE}30.8 $\pm$ 0.3 & \cellcolor[HTML]{8ED2B1}30.9 $\pm$ 0.1 & \cellcolor[HTML]{79C9A2}31.1 $\pm$ 0.2
\end{tabular}
\end{adjustbox}
\caption{Result of Image$ \rightarrow$ Text and Text $\rightarrow$ Image Recall@1 with different gradient combinations on two steps VSE++ training with ResNet152.}
\label{table:vseppresnet}
\end{table*}

\begin{table*}[t]
\centering
\begin{adjustbox}{width=1.9\columnwidth}
\begin{tabular}{c|ccc|ccc}
\hline
\multicolumn{7}{c}{VSE++ (ViT-base-patch16)}\\
\hline
& \multicolumn{3}{c|}{Image$ \rightarrow$ Text} 
& \multicolumn{3}{c}{Text $\rightarrow$ Image} \\
\hline
& $T^{con}$ & $T^{nca}$ & $T^{cir}$ & $T^{con}$ & $T^{nca}$ & $T^{cir}$ \\
\hline
$P^{con}$    & \cellcolor[HTML]{FFFFFF}37.6 $\pm$ 0.2 & \cellcolor[HTML]{A9C8FA}38.8 $\pm$ 0.1 & \cellcolor[HTML]{EAF2FE}37.9 $\pm$ 0.1 & \cellcolor[HTML]{FFFFFF}26.4 $\pm$ 0.2 & \cellcolor[HTML]{E7F0FE}26.6 $\pm$ 0.1 & \cellcolor[HTML]{FCFDFF}26.4 $\pm$ 0.1 \\
$P^{lin}$    & \cellcolor[HTML]{F8FBFF}37.7 $\pm$ 0.3 & \cellcolor[HTML]{CADDFC}38.3 $\pm$ 0.3 & \cellcolor[HTML]{9BBEFA}39.0 $\pm$ 0.2 & \cellcolor[HTML]{AAC9FB}27.0 $\pm$ 0.2 & \cellcolor[HTML]{9ABEFA}27.1 $\pm$ 0.0 & \cellcolor[HTML]{AFCCFB}27.0 $\pm$ 0.3 \\
$P^{sig}$    & \cellcolor[HTML]{C5DAFC}38.4 $\pm$ 0.5 & \cellcolor[HTML]{4F8DF5}40.0 $\pm$ 0.4 & \cellcolor[HTML]{75A6F7}39.5 $\pm$ 0.8 & \cellcolor[HTML]{8FB7F9}27.2 $\pm$ 0.0 & \cellcolor[HTML]{649BF6}27.5 $\pm$ 0.1 & \cellcolor[HTML]{77A7F8}27.4 $\pm$ 0.3 \\
$P^{lin-ms}$ & \cellcolor[HTML]{DAE7FD}38.1 $\pm$ 0.3 & \cellcolor[HTML]{8FB7F9}39.1 $\pm$ 0.3 & \cellcolor[HTML]{9ABEFA}39.0 $\pm$ 0.4 & \cellcolor[HTML]{93B9F9}27.2 $\pm$ 0.1 & \cellcolor[HTML]{86B1F8}27.3 $\pm$ 0.1 & \cellcolor[HTML]{B1CDFB}27.0 $\pm$ 0.5 \\
$P^{sig-ms}$ & \cellcolor[HTML]{518FF5}39.9 $\pm$ 0.3 & \cellcolor[HTML]{4285F4}40.1 $\pm$ 0.3 & \cellcolor[HTML]{689EF7}39.6 $\pm$ 0.6 & \cellcolor[HTML]{4285F4}27.7 $\pm$ 0.1 & \cellcolor[HTML]{5C96F6}27.6 $\pm$ 0.1 & \cellcolor[HTML]{7BAAF8}27.3 $\pm$ 0.2 \\
\hline
\multicolumn{7}{c}{VSE++ (ViT-base-patch16, fine-tuned)}\\
\hline
$P^{con}$    & \cellcolor[HTML]{FFFFFF}48.2 $\pm$ 0.5 & \cellcolor[HTML]{B5D0FB}49.6 $\pm$ 0.6 & \cellcolor[HTML]{E8F1FE}48.6 $\pm$ 0.9 & \cellcolor[HTML]{F2F7FF}36.5 $\pm$ 0.3 & \cellcolor[HTML]{B7D1FB}37.0 $\pm$ 0.2 & \cellcolor[HTML]{D5E4FD}36.7 $\pm$ 0.3 \\
$P^{lin}$    & \cellcolor[HTML]{F8FBFF}48.3 $\pm$ 0.3 & \cellcolor[HTML]{C0D7FC}49.4 $\pm$ 0.3 & \cellcolor[HTML]{C6DAFC}49.3 $\pm$ 0.3 & \cellcolor[HTML]{FFFFFF}36.4 $\pm$ 0.1 & \cellcolor[HTML]{9DC0FA}37.2 $\pm$ 0.3 & \cellcolor[HTML]{89B3F9}37.4 $\pm$ 0.2 \\
$P^{sig}$    & \cellcolor[HTML]{CADDFC}49.2 $\pm$ 0.6 & \cellcolor[HTML]{6EA1F7}50.9 $\pm$ 0.2 & \cellcolor[HTML]{629AF6}51.1 $\pm$ 0.4 & \cellcolor[HTML]{9EC0FA}37.2 $\pm$ 0.3 & \cellcolor[HTML]{4C8BF5}37.9 $\pm$ 0.3 & \cellcolor[HTML]{4285F4}37.9 $\pm$ 0.2 \\
$P^{lin-ms}$ & \cellcolor[HTML]{D6E5FD}48.9 $\pm$ 0.4 & \cellcolor[HTML]{93B9F9}50.2 $\pm$ 0.1 & \cellcolor[HTML]{AFCBFB}49.7 $\pm$ 0.2 & \cellcolor[HTML]{D1E2FD}36.8 $\pm$ 0.2 & \cellcolor[HTML]{80ADF8}37.4 $\pm$ 0.1 & \cellcolor[HTML]{70A3F7}37.6 $\pm$ 0.4 \\
$P^{sig-ms}$ & \cellcolor[HTML]{8AB4F9}50.4 $\pm$ 0.8 & \cellcolor[HTML]{78A8F8}50.7 $\pm$ 0.5 & \cellcolor[HTML]{4285F4}51.7 $\pm$ 0.2 & \cellcolor[HTML]{82AEF8}37.4 $\pm$ 0.4 & \cellcolor[HTML]{93B9F9}37.3 $\pm$ 0.1 & \cellcolor[HTML]{4587F5}37.9 $\pm$ 0.2 
\end{tabular}
\end{adjustbox}
\caption{Result of Image$ \rightarrow$ Text and Text $\rightarrow$ Image Recall@1 with different gradient combinations on two steps VSE++ training with ViT.}
\label{table:vseppvit}
\end{table*}

\section{Experiments}
\label{sec:experiments}
\subsection{Settings}
We run a set of experiments on the MS-COCO~\cite{chen2015microsoft} and Flickr~\cite{flickr} dataset. All experiments are run on the PyTorch platform~\cite{pytorch} with Nvidia Tesla V100 32GB GPU. We directly replace the loss module with gradient objective in three open source works: VSE++~\cite{faghri2018vse++}, VSE$\infty$~\cite{chen2021vseinfty} and X-VLM~\cite{xvlm} and keep all other training settings the same as their original work. We test all possible gradient objectives formed by the combination of the triplet weights and pair weights in Section~\ref{sec:w_triplet} and~\ref{sec:w_pair} for these three works. Each objective is run for 3 times to remove the effect caused by the randomness coming from the random sampling of the batch and random initialization of the mapping layers to joint space. We report two common retrieval results, image to text retrieval and text to image retrieval, with mean and standard deviation of Recall@1 as metric for both datasets. We show MS-COCO 5K test result in the main paper and Flickr 1k test result in the appendix. 

\subsection{Validation on Gradient Method}
In Tabel~\ref{table:verify_short}, we show the results from origin VSE++ and VSE$\infty$ work trained with triplet loss and the results implemented with the equivalent gradient methods with combination of $T^{con}$ and $P^{con}$. For VSE++ method, we re-implement the experiment ``ResNet152, fine-tune" result, denoted as ``VSE++(R152,FT) ours". For VSE$\infty$, we re-implement the experiment with pre-extracted object features (BUTD feature) and the Grid features with a pretrained model on Instagram (WSL)~\cite{Mahajan_2018_ECCV}, denoted as ``VSE$\infty$(BUTD) ours" and ``VSE$\infty$(WSL) ours". The re-implemented results are almost the same as originally reported numbers, validating our gradient objective with combination of $T^{con}$ and $P^{con}$ has equivalent effect to the triplet loss.

\subsection{Results on VSE++ and VSE$\infty$}
\textbf{VSE++} divides the training into two steps. The first step is to freeze the image extractor backbone and train the text extractor and the mapping layers to joint space. In the second step, all parameters of the image and text extractors and the mapping layers are included in the training. We re-implement the original experiments VSE++ (ResNet152) and VSE++ (ResNet152, fine-tuned) for these two steps and replace the triplet loss function with all possible gradient objectives. In addition, we run the same experiments with ViT~\cite{dosovitskiy2020vit}(ViT-base-patch16) which has been popularly used in vision language tasks to compare the performance of gradient objective on different models.

In Table~\ref{table:vseppresnet} and~\ref{table:vseppvit}, the pair weights $P^{lin}$, $P^{sig}$, $P^{lin-ms}$, $P^{sig-ms}$ show clear improvement in Recall@1 over the baseline pair weight $P^{con}$. In addition to pair weight, triplet weights $T^{nca}$, $T^{cir}$ help pair weight continue to improve the Recall@1 results in the fine-tuning step. 

Besides, all DML loss functions mentioned in Table~\ref{table:methods} perform better than triplet loss in both steps. In the fine-tuning step, we find the best loss function is MS loss($P^{sig-ms}$,$T^{con}$). But it is still sub-optimal when we combine triplet weight $T^{nca}$ or $T^{cir}$ with pair weight $P^{sig-ms}$, demonstrating the advantage of exploring the gradient space with GOAL. 

\begin{table*}[t]
\centering
\begin{adjustbox}{width=1.9\columnwidth}
\begin{tabular}{c|ccc|ccc}
\hline
\multicolumn{7}{c}{VSE$\infty$(BUTD)}\\
\hline
& \multicolumn{3}{c|}{Image$ \rightarrow$ Text} 
& \multicolumn{3}{c}{Text $\rightarrow$ Image} \\
\hline
& $T^{con}$ & $T^{nca}$ & $T^{cir}$ & $T^{con}$ & $T^{nca}$ & $T^{cir}$ \\
\hline
$P^{con}$    & \cellcolor[HTML]{FBEAE9}58.9 $\pm$ 0.7 & \cellcolor[HTML]{F0B0AA}61.2 $\pm$ 0.7 & \cellcolor[HTML]{F5C8C4}60.3 $\pm$ 0.2 & \cellcolor[HTML]{F9DDDB}43.1 $\pm$ 0.0 & \cellcolor[HTML]{F8D9D7}43.2 $\pm$ 0.3 & \cellcolor[HTML]{FFFFFF}42.5 $\pm$ 0.1 \\
$P^{lin}$    & \cellcolor[HTML]{FFFFFF}58.1 $\pm$ 0.2 & \cellcolor[HTML]{F3BDB8}60.7 $\pm$ 0.1 & \cellcolor[HTML]{F6CBC8}60.1 $\pm$ 0.2 & \cellcolor[HTML]{F9E0DD}43.1 $\pm$ 0.3 & \cellcolor[HTML]{F6CECA}43.4 $\pm$ 0.2 & \cellcolor[HTML]{FAE2E0}43.0 $\pm$ 0.1 \\
$P^{sig}$    & \cellcolor[HTML]{F7D4D1}59.8 $\pm$ 0.6 & \cellcolor[HTML]{EC9B94}62.0 $\pm$ 0.4 & \cellcolor[HTML]{EC9A93}62.0 $\pm$ 0.4 & \cellcolor[HTML]{F5C7C3}43.5 $\pm$ 0.2 & \cellcolor[HTML]{F1B6B1}43.9 $\pm$ 0.1 & \cellcolor[HTML]{F1B6B1}43.8 $\pm$ 0.2 \\
$P^{lin-ms}$ & \cellcolor[HTML]{F6CECB}60.0 $\pm$ 0.1 & \cellcolor[HTML]{EEA39D}61.7 $\pm$ 0.2 & \cellcolor[HTML]{F0B0AA}61.2 $\pm$ 0.2 & \cellcolor[HTML]{F4C5C1}43.6 $\pm$ 0.2 & \cellcolor[HTML]{F1B3AE}43.9 $\pm$ 0.2 & \cellcolor[HTML]{F6CECB}43.4 $\pm$ 0.2 \\
$P^{sig-ms}$ & \cellcolor[HTML]{ED9F98}61.8 $\pm$ 0.2 & \cellcolor[HTML]{E77E75}63.1 $\pm$ 0.2 & \cellcolor[HTML]{E67C73}63.2 $\pm$ 0.3 & \cellcolor[HTML]{EA8F88}44.6 $\pm$ 0.2 & \cellcolor[HTML]{E8837B}44.8 $\pm$ 0.1 & \cellcolor[HTML]{E67C73}44.9 $\pm$ 0.1 \\
\hline
\multicolumn{7}{c}{VSE$\infty$(WSL)}\\
\hline
$P^{con}$    & \cellcolor[HTML]{FFFFFF}66.2 $\pm$ 0.2 & \cellcolor[HTML]{F9DDDA}67.6 $\pm$ 0.4 & \cellcolor[HTML]{FBE7E6}67.2 $\pm$ 0.4 & \cellcolor[HTML]{F5CBC7}51.6 $\pm$ 0.3 & \cellcolor[HTML]{F6D0CD}51.4 $\pm$ 0.2 & \cellcolor[HTML]{FFFFFF}49.9 $\pm$ 0.1 \\
$P^{lin}$    & \cellcolor[HTML]{FCEFED}66.9 $\pm$ 0.7 & \cellcolor[HTML]{F5C7C4}68.5 $\pm$ 0.4 & \cellcolor[HTML]{F5CBC7}68.4 $\pm$ 0.5 & \cellcolor[HTML]{F0AFAA}52.5 $\pm$ 0.1 & \cellcolor[HTML]{EEA6A0}52.8 $\pm$ 0.2 & \cellcolor[HTML]{F0B0AB}52.5 $\pm$ 0.2 \\
$P^{sig}$    & \cellcolor[HTML]{F6CECA}68.2 $\pm$ 0.6 & \cellcolor[HTML]{EFAAA4}69.7 $\pm$ 0.2 & \cellcolor[HTML]{ED9E97}70.2 $\pm$ 0.5 & \cellcolor[HTML]{EDA19B}53.0 $\pm$ 0.1 & \cellcolor[HTML]{ED9D96}53.1 $\pm$ 0.2 & \cellcolor[HTML]{EEA49D}52.9 $\pm$ 0.2 \\
$P^{lin-ms}$ & \cellcolor[HTML]{F8D8D6}67.8 $\pm$ 0.1 & \cellcolor[HTML]{EFABA5}69.7 $\pm$ 0.2 & \cellcolor[HTML]{F0AFAA}69.5 $\pm$ 0.1 & \cellcolor[HTML]{EFA8A2}52.8 $\pm$ 0.2 & \cellcolor[HTML]{EC9790}53.3 $\pm$ 0.2 & \cellcolor[HTML]{EFABA5}52.7 $\pm$ 0.1 \\
$P^{sig-ms}$ & \cellcolor[HTML]{EC9A93}70.3 $\pm$ 0.2 & \cellcolor[HTML]{E67C73}71.5 $\pm$ 0.4 & \cellcolor[HTML]{E77E75}71.4 $\pm$ 0.5 & \cellcolor[HTML]{E8857D}53.9 $\pm$ 0.4 & \cellcolor[HTML]{E67C73}54.2 $\pm$ 0.0 & \cellcolor[HTML]{EA8D85}53.6 $\pm$ 0.6
\end{tabular}
\end{adjustbox}
\caption{Result of Image$ \rightarrow$ Text and Text $\rightarrow$ Image Recall@1 with different gradient combinations on VSE$\infty$(BUTD) and VSE$\infty$(WSL).}
\label{table:vseinfty}
\end{table*}
\begin{table*}[t]
\centering
\begin{adjustbox}{width=2\columnwidth}
\begin{tabular}{l|cc|ccc|ccc|ccc|ccc}
& & & \multicolumn{6}{c|}{MS-COCO 5K test} & \multicolumn{6}{c}{Flickr 1K test} \\
\hline
Tasks & & & \multicolumn{3}{c|}{Image$ \rightarrow$ Text} & \multicolumn{3}{c|}{Text$ \rightarrow$ Image} & \multicolumn{3}{c|}{Image$ \rightarrow$ Text} & \multicolumn{3}{c}{Text$ \rightarrow$ Image} \\
\hline
Method & Pre-train & Data size & R@1 & R@5 & R@10 & R@1 & R@5 & R@10 & R@1 & R@5 & R@10 & R@1 & R@5 & R@10 \\
\hline
VSE++(R152, FT)~\cite{faghri2018vse++}
& \xmark & - 
& 41.3 & - & 81.2 & 30.3 & - & 72.4
& 52.9 & - & 87.2 & 39.6 & - & 79.5 \\
VSE++(R152, FT) \textbf{ours}
& \xmark & - 
& 44.3 & 73.3 & 83.7 & 31.0 & 60.2 & 72.5
& 57.0 & 82.6 & 89.2 & 42.4 & 72.4 & 81.0\\%nca-sig-cos
SCAN~\cite{scan}
& \xmark & - 
& 50.4 & 82.2 & 90.0 & 38.6 & 69.3 & 80.4 
& 67.4 & 90.3 & 95.8 & 48.6 & 77.7 & 85.2 \\
VSRN~\cite{li2019visual}
& \xmark & - 
& 53.0 & 81.1 & 89.4 & 40.5 & 70.6 & 81.1 
& 71.3 & 90.6 & 96.0 & 54.7 & 81.8 & 88.2 \\
VSE$\infty$(BUTD)~\cite{chen2021vseinfty}
& \xmark & - 
& 58.3 & 85.3 & - & 42.4 & 72.7 & - 
& 81.7 & 95.4 & 97.6 & 61.4 & 85.9 & 91.5 \\
VSE$\infty$(BUTD) \textbf{ours}
& \xmark & - 
& 63.2 & 87.2 & 93.0 & 44.4 & 74.2 & 83.9 
& 82.3 & 95.8 & 98.4 & 64.0 & 87.5 & 92.7 \\
VSE$\infty$(WSL)~\cite{chen2021vseinfty}
& \xmark & - 
& 66.4 & 89.3 & - & 51.6 & 79.3 & - 
& 88.4 & 98.3 & 99.5 & 74.2 & 93.7 & 96.8\\
VSE$\infty$(WSL) \textbf{ours}
& \xmark & - 
& 71.9 & 92.0 & 95.9 & 53.7 & 80.6 & 88.4 
& 90.6 & 99.2 & 99.6 & 76.7 & 94.6 & 97.3\\
\hline
VinVL~\cite{zhang2021vinvl}
& \cmark & 5.6M 
& 75.4 & 92.9 & 96.2 & 58.8 & 83.5 & 90.3 
& - & - & - & - & - & - \\
% CVSE~\cite{wang2020consensus}
% & \xmark & 78.6 & 95.0 & 97.5 & 66.3 & 91.8 & \textbf{96.3} \\
ALBEF~\cite{ALBEF}
& \cmark & 14M 
& 77.6 & 94.3 & 97.2 & 60.7 & 84.3 & 90.5
& 95.9 & \textbf{99.8} & 100.0 & 85.6 & \textbf{97.5} & 98.9 \\
X-VLM~\cite{xvlm}
& \cmark & 4M 
& 80.4 & 95.5 & \textbf{98.2} & 63.1 & 85.7 & \textbf{91.6}
& 96.8 & 99.8 & 100.0 & 86.1 & 97.4 & 98.7 \\
X-VLM \textbf{ours}
& \cmark & 4M 
& \textbf{81.4} & \textbf{95.6} & 97.9 & \textbf{63.6} & \textbf{86.0} & 91.5
& \textbf{97.0} & 99.6 & \textbf{100.0} & \textbf{86.3} & 97.4 & \textbf{99.0} \\
\hline
\end{tabular}
\end{adjustbox}
% \caption{Image-text retrieval results on MS-COCO 5K test with variants of VSE method and VLP downstream training}
\caption{Our state-of-the art image-text retrieval results on MS-COCO 5K and Flickr 1K test using the novel loss function designed with the proposed GOAL framework.}
\label{table:SOTA}
\end{table*}

% TODO: 
% (1) ALBEF, ALIGN (2) add # pretrain-images (3) add VSE++/VSE_inf with gradient losses (noted at suffixes) 

\textbf{VSE$\infty$} We re-implement two training setups in VSE$\infty$. Both setups use BERT-base~\cite{devlin-etal-2019-bert} as the text feature extractor. For the image feature, one uses the pre-extracted object features (BUTD feature) and another uses the grid features with a pretrained model on Instagram (WSL feature)~\cite{Mahajan_2018_ECCV}. A learned Generalized Pooling Operator(GPO) aggregates and projects the image and text feature vectors independently into the joint embedding space to further compute the loss. Still, we only replace the triplet loss function used in the training with gradient objectives.

Table~\ref{table:vseinfty} shows similar improvement pattern as shown in the result of VSE++, verifying our GOAL is general to different of VSE methods.

\subsection{State-of-the-Art Results}
Finally, we compare two sets of state-of-the-art approaches on MS-COCO 5K test and Flickr 1K test. One set is VSE related and another set is VLP related. In Table~\ref{table:SOTA},  We first show our best improved result of VSE++ and VSE$\infty$ method, denoted as ``VSE++(R152, FT) ours",  ``VSE$\infty$(BUTD) ours" and ``VSE$\infty$(BUTD) ours", which are trained with combination of ($T^{cir}$, $P^{sig-ms}$). 
In MS-COCO 5K test, the gain of Image$ \rightarrow$ Text R@1 and Text$ \rightarrow$ Image R@1 on VSE++(R152, FT) is 3$\%$, 0.7$\%$, on VSE$\infty$(BUTD) is 4.9$\%$, 2.0$\%$ and on VSE$\infty$(WSL) is 5.5$\%$, 2.1$\%$. 
In Flickr 1K test, the gain of Image$ \rightarrow$ Text R@1 and Text$ \rightarrow$ Image R@1 on VSE++(R152, FT) is 4.1$\%$, 2.8$\%$, on VSE$\infty$(BUTD) is 0.6$\%$, 2.6$\%$ and on VSE$\infty$(WSL) is 2.2$\%$, 2.5$\%$.

In addition, we apply the same gradient objective in the latest state-of-the-art approach X-VLM~\cite{xvlm} with replacement of its contrastive loss item in the downstream fine-tuning. The result is denoted as ``X-VLM ours". We continue to push the boundary of state-of-the-art result on MS-COCO 5K test and Flickr 1K test.

\section{Conclusion}
We provide a new framework GOAL to train image-text matching tasks with a combination of gradient components dissected from deep metric learning loss functions. In practice, the proposed gradient objectives can be easily applied as a drop-in replacement to training with loss functions. Extensive experiments on exhaustive combinations of triplet weights and pair weights demonstrate both triplet weights and pair weights have individual impact on the retrieval performance and generally the combination of  $T^{cir}$, $P^{sig-ms}$ achieve the best performance on image-text retrieval. This framework helps find better gradient objectives which have never been explored for this domain and provides consistent retrieval improvement on multiple established methods, including achieving new state-of-the-art results. 

{\small
\bibliographystyle{ieee_fullname}
\bibliography{egbib}

\begin{thebibliography}{10}\itemsep=-1pt

\bibitem{butd_anderson2018bottom}
Peter Anderson, Xiaodong He, Chris Buehler, Damien Teney, Mark Johnson, Stephen
  Gould, and Lei Zhang.
\newblock Bottom-up and top-down attention for image captioning and visual
  question answering.
\newblock In {\em Proceedings of the IEEE conference on computer vision and
  pattern recognition}, pages 6077--6086, 2018.

\bibitem{chen2021vseinfty}
Jiacheng Chen, Hexiang Hu, Hao Wu, Yuning Jiang, and Changhu Wang.
\newblock Learning the best pooling strategy for visual semantic embedding.
\newblock In {\em IEEE Conference on Computer Vision and Pattern Recognition
  (CVPR)}, 2021.

\bibitem{chen2015microsoft}
Xinlei Chen, Hao Fang, Tsung-Yi Lin, Ramakrishna Vedantam, Saurabh Gupta, Piotr
  Doll{\'a}r, and C~Lawrence Zitnick.
\newblock Microsoft coco captions: Data collection and evaluation server.
\newblock {\em arXiv preprint arXiv:1504.00325}, 2015.

\bibitem{chen2020uniter}
Yen-Chun Chen, Linjie Li, Licheng Yu, Ahmed~El Kholy, Faisal Ahmed, Zhe Gan, Yu
  Cheng, and Jingjing Liu.
\newblock Uniter: Universal image-text representation learning.
\newblock In {\em ECCV}, 2020.

\bibitem{devlin-etal-2019-bert}
Jacob Devlin, Ming-Wei Chang, Kenton Lee, and Kristina Toutanova.
\newblock {BERT}: Pre-training of deep bidirectional transformers for language
  understanding.
\newblock In {\em Proceedings of the 2019 Conference of the North {A}merican
  Chapter of the Association for Computational Linguistics: Human Language
  Technologies, Volume 1 (Long and Short Papers)}, pages 4171--4186,
  Minneapolis, Minnesota, June 2019. Association for Computational Linguistics.

\bibitem{dosovitskiy2020vit}
Alexey Dosovitskiy, Lucas Beyer, Alexander Kolesnikov, Dirk Weissenborn,
  Xiaohua Zhai, Thomas Unterthiner, Mostafa Dehghani, Matthias Minderer, Georg
  Heigold, Sylvain Gelly, et~al.
\newblock An image is worth 16x16 words: Transformers for image recognition at
  scale.
\newblock {\em arXiv preprint arXiv:2010.11929}, 2020.

\bibitem{faghri2018vse++}
Fartash Faghri, David~J Fleet, Jamie~Ryan Kiros, and Sanja Fidler.
\newblock Vse++: Improving visual-semantic embeddings with hard negatives.
\newblock In {\em Proceedings of the British Machine Vision Conference
  ({BMVC})}, 2018.

\bibitem{frome2013devise}
Andrea Frome, Greg~S Corrado, Jon Shlens, Samy Bengio, Jeff Dean, Marc'Aurelio
  Ranzato, and Tomas Mikolov.
\newblock Devise: A deep visual-semantic embedding model.
\newblock {\em Advances in neural information processing systems}, 26, 2013.

\bibitem{nca}
Jacob Goldberger, Geoffrey~E Hinton, Sam~T. Roweis, and Ruslan~R Salakhutdinov.
\newblock Neighbourhood components analysis.
\newblock In L.~K. Saul, Y. Weiss, and L. Bottou, editors, {\em Advances in
  Neural Information Processing Systems 17}, pages 513--520. MIT Press, 2005.

\bibitem{hadsell2006dimensionality}
Raia Hadsell, Sumit Chopra, and Yann LeCun.
\newblock Dimensionality reduction by learning an invariant mapping.
\newblock In {\em Proc. IEEE Conference on Computer Vision and Pattern
  Recognition (CVPR)}, volume~2, pages 1735--1742. IEEE, 2006.

\bibitem{resnet}
Kaiming He, Xiangyu Zhang, Shaoqing Ren, and Jian Sun.
\newblock Deep residual learning for image recognition.
\newblock In {\em Proc. IEEE Conference on Computer Vision and Pattern
  Recognition (CVPR)}, June 2016.

\bibitem{hochreiter1997lstm}
Sepp Hochreiter and J{\"u}rgen Schmidhuber.
\newblock Long short-term memory.
\newblock {\em Neural computation}, 9(8):1735--1780, 1997.

\bibitem{hoffer2015deep}
Elad Hoffer and Nir Ailon.
\newblock Deep metric learning using triplet network.
\newblock In {\em International Workshop on Similarity-Based Pattern
  Recognition}, pages 84--92. Springer, 2015.

\bibitem{huang2018learning}
Yan Huang, Qi Wu, Chunfeng Song, and Liang Wang.
\newblock Learning semantic concepts and order for image and sentence matching.
\newblock In {\em Proceedings of the IEEE Conference on Computer Vision and
  Pattern Recognition}, pages 6163--6171, 2018.

\bibitem{kiros2014unifying}
Ryan Kiros, Ruslan Salakhutdinov, and Richard~S Zemel.
\newblock Unifying visual-semantic embeddings with multimodal neural language
  models.
\newblock {\em arXiv preprint arXiv:1411.2539}, 2014.

\bibitem{CAR196}
Jonathan Krause, Michael Stark, Jia Deng, and Li Fei-Fei.
\newblock 3d object representations for fine-grained categorization.
\newblock In {\em 4th International IEEE Workshop on 3D Representation and
  Recognition (3dRR-13)}, Sydney, Australia, 2013.

\bibitem{scan}
Kuang-Huei Lee, Xi Chen, Gang Hua, Houdong Hu, and Xiaodong He.
\newblock Stacked cross attention for image-text matching.
\newblock In {\em Proceedings of the European Conference on Computer Vision
  (ECCV)}, pages 201--216, 2018.

\bibitem{ALBEF}
Junnan Li, Ramprasaath~R. Selvaraju, Akhilesh~Deepak Gotmare, Shafiq Joty,
  Caiming Xiong, and Steven Hoi.
\newblock Align before fuse: Vision and language representation learning with
  momentum distillation.
\newblock In {\em NeurIPS}, 2021.

\bibitem{li2019visual}
Kunpeng Li, Yulun Zhang, Kai Li, Yuanyuan Li, and Yun Fu.
\newblock Visual semantic reasoning for image-text matching.
\newblock In {\em Proceedings of the IEEE/CVF International conference on
  computer vision}, pages 4654--4662, 2019.

\bibitem{li2020oscar}
Xiujun Li, Xi Yin, Chunyuan Li, Pengchuan Zhang, Xiaowei Hu, Lei Zhang, Lijuan
  Wang, Houdong Hu, Li Dong, Furu Wei, et~al.
\newblock Oscar: Object-semantics aligned pre-training for vision-language
  tasks.
\newblock In {\em European Conference on Computer Vision}, pages 121--137.
  Springer, 2020.

\bibitem{liu2019roberta}
Yinhan Liu, Myle Ott, Naman Goyal, Jingfei Du, Mandar Joshi, Danqi Chen, Omer
  Levy, Mike Lewis, Luke Zettlemoyer, and Veselin Stoyanov.
\newblock Roberta: A robustly optimized bert pretraining approach.
\newblock {\em arXiv preprint arXiv:1907.11692}, 2019.

\bibitem{ICR}
Ziwei Liu, Ping Luo, Shi Qiu, Xiaogang Wang, and Xiaoou Tang.
\newblock Deepfashion: Powering robust clothes recognition and retrieval with
  rich annotations.
\newblock In {\em Proceedings of IEEE Conference on Computer Vision and Pattern
  Recognition (CVPR)}, June 2016.

\bibitem{lu2019vilbert}
Jiasen Lu, Dhruv Batra, Devi Parikh, and Stefan Lee.
\newblock Vilbert: Pretraining task-agnostic visiolinguistic representations
  for vision-and-language tasks.
\newblock {\em Advances in neural information processing systems}, 32, 2019.

\bibitem{Mahajan_2018_ECCV}
Dhruv Mahajan, Ross Girshick, Vignesh Ramanathan, Kaiming He, Manohar Paluri,
  Yixuan Li, Ashwin Bharambe, and Laurens van~der Maaten.
\newblock Exploring the limits of weakly supervised pretraining.
\newblock In {\em Proceedings of the European Conference on Computer Vision
  (ECCV)}, September 2018.

\bibitem{pytorch}
Adam Paszke, Sam Gross, Soumith Chintala, Gregory Chanan, Edward Yang, Zachary
  DeVito, Zeming Lin, Alban Desmaison, Luca Antiga, and Adam Lerer.
\newblock Automatic differentiation in pytorch.
\newblock In {\em NIPS-W}, 2017.

\bibitem{facenet}
Florian Schroff, Dmitry Kalenichenko, and James Philbin.
\newblock Facenet: A unified embedding for face recognition and clustering.
\newblock In {\em Proc. IEEE Conference on Computer Vision and Pattern
  Recognition (CVPR)}, June 2015.

\bibitem{Npairs}
Kihyuk Sohn.
\newblock Improved deep metric learning with multi-class n-pair loss objective.
\newblock In {\em Advances in Neural Information Processing Systems}, pages
  1857--1865, 2016.

\bibitem{SOP}
Hyun~Oh Song, Yu Xiang, Stefanie Jegelka, and Silvio Savarese.
\newblock Deep metric learning via lifted structured feature embedding.
\newblock In {\em Proc. IEEE Conference on Computer Vision and Pattern
  Recognition (CVPR)}, 2016.

\bibitem{Sun_2020_CVPR}
Yifan Sun, Changmao Cheng, Yuhan Zhang, Chi Zhang, Liang Zheng, Zhongdao Wang,
  and Yichen Wei.
\newblock Circle loss: A unified perspective of pair similarity optimization.
\newblock In {\em IEEE/CVF Conference on Computer Vision and Pattern
  Recognition (CVPR)}, June 2020.

\bibitem{tan2019lxmert}
Hao Tan and Mohit Bansal.
\newblock Lxmert: Learning cross-modality encoder representations from
  transformers.
\newblock {\em arXiv preprint arXiv:1908.07490}, 2019.

\bibitem{vaswani2017attention}
Ashish Vaswani, Noam Shazeer, Niki Parmar, Jakob Uszkoreit, Llion Jones,
  Aidan~N Gomez, {\L}ukasz Kaiser, and Illia Polosukhin.
\newblock Attention is all you need.
\newblock {\em Advances in neural information processing systems}, 30, 2017.

\bibitem{wang2020consensus}
Haoran Wang, Ying Zhang, Zhong Ji, Yanwei Pang, and Lin Ma.
\newblock Consensus-aware visual-semantic embedding for image-text matching.
\newblock In {\em European Conference on Computer Vision}, pages 18--34.
  Springer, 2020.

\bibitem{wang2019multi}
Xun Wang, Xintong Han, Weilin Huang, Dengke Dong, and Matthew~R Scott.
\newblock Multi-similarity loss with general pair weighting for deep metric
  learning.
\newblock In {\em Proceedings of the IEEE Conference on Computer Vision and
  Pattern Recognition}, pages 5022--5030, 2019.

\bibitem{CUB200}
P. Welinder, S. Branson, T. Mita, C. Wah, F. Schroff, S. Belongie, and P.
  Perona.
\newblock {Caltech-UCSD Birds 200}.
\newblock Technical Report CNS-TR-2010-001, California Institute of Technology,
  2010.

\bibitem{wu2019unified}
Hao Wu, Jiayuan Mao, Yufeng Zhang, Yuning Jiang, Lei Li, Weiwei Sun, and
  Wei-Ying Ma.
\newblock Unified visual-semantic embeddings: Bridging vision and language with
  structured meaning representations.
\newblock In {\em Proceedings of the IEEE/CVF Conference on Computer Vision and
  Pattern Recognition}, pages 6609--6618, 2019.

\bibitem{Xuan_2020_ECCV}
Hong Xuan, Abby Stylianou, Xiaotong Liu, and Robert Pless.
\newblock Hard negative examples are hard, but useful.
\newblock In {\em The European Conference on Computer Vision (ECCV)}, September
  2020.

\bibitem{yi2014deep}
Dong Yi, Zhen Lei, Shengcai Liao, and Stan~Z Li.
\newblock Deep metric learning for person re-identification.
\newblock In {\em 2014 22nd International Conference on Pattern Recognition},
  pages 34--39. IEEE, 2014.

\bibitem{flickr}
Peter Young, Alice Lai, Micah Hodosh, and Julia Hockenmaier.
\newblock From image descriptions to visual denotations: New similarity metrics
  for semantic inference over event descriptions.
\newblock {\em Transactions of the Association for Computational Linguistics},
  2:67--78, 2014.

\bibitem{yu2020gradient}
Tianhe Yu, Saurabh Kumar, Abhishek Gupta, Sergey Levine, Karol Hausman, and
  Chelsea Finn.
\newblock Gradient surgery for multi-task learning.
\newblock {\em arXiv preprint arXiv:2001.06782}, 2020.

\bibitem{xvlm}
Yan Zeng, Xinsong Zhang, and Hang Li.
\newblock Multi-grained vision language pre-training: Aligning texts with
  visual concepts.
\newblock {\em arXiv preprint arXiv:2111.08276}, 2021.

\bibitem{zhang2021vinvl}
Pengchuan Zhang, Xiujun Li, Xiaowei Hu, Jianwei Yang, Lei Zhang, Lijuan Wang,
  Yejin Choi, and Jianfeng Gao.
\newblock Vinvl: Revisiting visual representations in vision-language models.
\newblock In {\em Proceedings of the IEEE/CVF Conference on Computer Vision and
  Pattern Recognition}, pages 5579--5588, 2021.

\bibitem{zhang2019p2sgrad}
Xiao Zhang, Rui Zhao, Junjie Yan, Mengya Gao, Yu Qiao, Xiaogang Wang, and
  Hongsheng Li.
\newblock P2sgrad: Refined gradients for optimizing deep face models.
\newblock In {\em Proceedings of the IEEE/CVF Conference on Computer Vision and
  Pattern Recognition}, pages 9906--9914, 2019.

\end{thebibliography}
}

\end{document}

\title{Appendix of Dissecting Deep Metric Learning Losses for Image-Text Retrieval}  % **** Enter the paper title here

\maketitle

\section{Validation on Gradient Method}
Table~\ref{table:verify} shows the results from origin VSE++ and VSE$\infty$ work trained with triplet loss and the results implemented with the equivalent gradient methods with combination of $T^{con}$ and $P^{con}$ as mentioned in Section 4.2.

\begin{table}[H] %[t]
\centering
\begin{adjustbox}{width=0.8\columnwidth}
\begin{tabular}{l|ccc|ccc}
\hline
& \multicolumn{3}{c|}{Image$ \rightarrow$ Text} 
& \multicolumn{3}{c}{Text $\rightarrow$ Image} \\
\hline
Method 
& R@1 & R@5 & R@10 & R@1 & R@5 & R@10 \\
\hline
VSE++(R152,FT) 
& 41.3 & - & 81.2 & 30.3 & - & 72.4 \\
VSE++(R152,FT) ours
& 41.0 $\pm$ 0.3 & 70.4 $\pm$ 0.4 & 81.3 $\pm$ 0.3 & 30.2 $\pm$ 0.1 & 60.1 $\pm$ 0.1 & 72.5 $\pm$ 0.1 \\
\hline
VSE$\infty$(BUTD)
& 58.3 & 85.3 & - & 42.4 & 72.7 & - \\
VSE$\infty$(BUTD) ours  
& 58.3 $\pm$ 0.7 & 85.5 $\pm$ 0.4 & 92.6 $\pm$ 0.0 & 43.1 $\pm$ 0.0 & 73.3 $\pm$ 0.1 & 83.4 $\pm$ 0.2 \\
\hline
VSE$\infty$(WSL)
& 66.4 & 89.3 & - & 51.6 & 79.3 & - \\
VSE$\infty$(WSL) ours  
& 66.2 $\pm$ 0.2 & 89.5 $\pm$ 0.2 & 94.8 $\pm$ 0.3 & 51.6 $\pm$ 0.3 & 79.3 $\pm$ 0.2 & 87.6 $\pm$ 0.2 \\
\hline
\end{tabular}
\end{adjustbox}
\caption{Results verification of the model trained with triplet loss function backward vs the model trained with gradient backward on three VSE methods}
\label{table:verify}
\end{table}

\section{Flickr 30K test result}
Similar Experiment on Flickr 30K as mentioned in Section 4.3.
\begin{table}[H]
\centering
\begin{adjustbox}{width=0.7\columnwidth}
\begin{tabular}{c|ccc|ccc}
\hline
\multicolumn{7}{c}{VSE++ (ResNet152, fine-tuned)}\\
\hline
& \multicolumn{3}{c|}{Image$ \rightarrow$ Text} 
& \multicolumn{3}{c}{Text $\rightarrow$ Image} \\
\hline
& $T^{con}$ & $T^{nca}$ & $T^{cir}$ & $T^{con}$ & $T^{nca}$ & $T^{cir}$ \\
\hline
$P^{con}$    & \cellcolor[HTML]{FFEFC4}55.0 $\pm$ 0.5 & \cellcolor[HTML]{FFEEBD}55.2 $\pm$ 0.3 & \cellcolor[HTML]{FFF3D2}54.7 $\pm$ 0.9 & \cellcolor[HTML]{FFEDB9}40.9 $\pm$ 0.7 & \cellcolor[HTML]{FFFFFF}39.6 $\pm$ 0.1 & \cellcolor[HTML]{FFF5D7}40.3 $\pm$ 0.2 \\
$P^{lin}$    & \cellcolor[HTML]{FFEAB1}55.4 $\pm$ 0.7 & \cellcolor[HTML]{FFF1CA}54.9 $\pm$ 1.1 & \cellcolor[HTML]{FFE9AA}55.6 $\pm$ 0.9 & \cellcolor[HTML]{FFE8A9}41.2 $\pm$ 0.5 & \cellcolor[HTML]{FFECB8}40.9 $\pm$ 0.1 & \cellcolor[HTML]{FFE59C}41.4 $\pm$ 0.4 \\
$P^{sig}$    & \cellcolor[HTML]{FFD970}56.8 $\pm$ 0.4 & \cellcolor[HTML]{FFD666}57.0 $\pm$ 0.5 & \cellcolor[HTML]{FFE089}56.3 $\pm$ 0.6 & \cellcolor[HTML]{FFE49A}41.5 $\pm$ 0.2 & \cellcolor[HTML]{FFD666}42.4 $\pm$ 0.5 & \cellcolor[HTML]{FFDB76}42.1 $\pm$ 0.4 \\
$P^{lin-ms}$ & \cellcolor[HTML]{FFF1C8}54.9 $\pm$ 0.6 & \cellcolor[HTML]{FFE18C}56.2 $\pm$ 0.7 & \cellcolor[HTML]{FFE7A4}55.7 $\pm$ 0.5 & \cellcolor[HTML]{FFF1C8}40.6 $\pm$ 0.8 & \cellcolor[HTML]{FFEAB0}41.1 $\pm$ 0.4 & \cellcolor[HTML]{FFEBB3}41.0 $\pm$ 0.3 \\
$P^{sig-ms}$ & \cellcolor[HTML]{FFFFFF}53.8 $\pm$ 0.7 & \cellcolor[HTML]{FFDE81}56.4 $\pm$ 1.3 & \cellcolor[HTML]{FFE7A6}55.7 $\pm$ 1.1 & \cellcolor[HTML]{FFF6DB}40.3 $\pm$ 0.4 & \cellcolor[HTML]{FFE9AD}41.1 $\pm$ 0.2 & \cellcolor[HTML]{FFF1C8}40.6 $\pm$ 0.2 \\
\hline
\multicolumn{7}{c}{VSE++ (ViT-base-patch16, fine-tuned)}\\
\hline
$P^{con}$    & \cellcolor[HTML]{FFFDF8}67.3 $\pm$ 0.9 & \cellcolor[HTML]{FFFFFF}67.1 $\pm$ 0.3 & \cellcolor[HTML]{FFF5D9}68.0 $\pm$ 0.6 & \cellcolor[HTML]{FFFFFF}52.8 $\pm$ 0.2 & \cellcolor[HTML]{FFF0C5}53.6 $\pm$ 0.4 & \cellcolor[HTML]{FFF7DF}53.3 $\pm$ 0.3 \\
$P^{lin}$    & \cellcolor[HTML]{FFFDF5}67.4 $\pm$ 0.6 & \cellcolor[HTML]{FFF3D2}68.1 $\pm$ 0.1 & \cellcolor[HTML]{FFF3D2}68.1 $\pm$ 0.4 & \cellcolor[HTML]{FFFFFF}52.8 $\pm$ 0.3 & \cellcolor[HTML]{FFF5D7}53.4 $\pm$ 0.2 & \cellcolor[HTML]{FFF2CD}53.5 $\pm$ 0.4 \\
$P^{sig}$    & \cellcolor[HTML]{FFE9AD}68.9 $\pm$ 0.2 & \cellcolor[HTML]{FFF0C5}68.4 $\pm$ 0.5 & \cellcolor[HTML]{FFEAB0}68.9 $\pm$ 0.2 & \cellcolor[HTML]{FFEFC3}53.6 $\pm$ 0.1 & \cellcolor[HTML]{FFDB79}54.6 $\pm$ 0.2 & \cellcolor[HTML]{FFE396}54.2 $\pm$ 0.2 \\
$P^{lin-ms}$ & \cellcolor[HTML]{FFEAAE}68.9 $\pm$ 0.8 & \cellcolor[HTML]{FFF0C7}68.4 $\pm$ 0.6 & \cellcolor[HTML]{FFE7A4}69.1 $\pm$ 1.0 & \cellcolor[HTML]{FFF9E8}53.1 $\pm$ 0.2 & \cellcolor[HTML]{FFF0C7}53.6 $\pm$ 0.4 & \cellcolor[HTML]{FFF3D1}53.4 $\pm$ 0.2 \\
$P^{sig-ms}$ & \cellcolor[HTML]{FFEBB3}68.8 $\pm$ 0.0 & \cellcolor[HTML]{FFD666}70.5 $\pm$ 1.4 & \cellcolor[HTML]{FFDB77}70.1 $\pm$ 1.6 & \cellcolor[HTML]{FFFBEE}53.1 $\pm$ 0.4 & \cellcolor[HTML]{FFD666}54.8 $\pm$ 0.2 & \cellcolor[HTML]{FFDF85}54.4 $\pm$ 0.3 \\
\hline
\end{tabular}
\end{adjustbox}
\caption{Result of Image$ \rightarrow$ Text and Text $\rightarrow$ Image Recall@1 on Flickr 30K test with different gradient combinations on two steps VSE++ training with ResNet152.}
\label{table:vseppf30k}
\end{table}

\begin{table}[H] %[t]
\centering
\begin{adjustbox}{width=0.7\columnwidth}
\begin{tabular}{c|ccc|ccc}
\hline
\multicolumn{7}{c}{VSE$\infty$(BUTD)}\\
\hline
& \multicolumn{3}{c|}{Image$ \rightarrow$ Text} 
& \multicolumn{3}{c}{Text $\rightarrow$ Image} \\
\hline
& $T^{con}$ & $T^{nca}$ & $T^{cir}$ & $T^{con}$ & $T^{nca}$ & $T^{cir}$ \\
\hline
$P^{con}$    & \cellcolor[HTML]{A3DAC0}81.1 $\pm$ 0.6 & \cellcolor[HTML]{73C79E}81.8 $\pm$ 0.8 & \cellcolor[HTML]{77C8A1}81.8 $\pm$ 0.9 & \cellcolor[HTML]{FFFFFF}62.1 $\pm$ 0.5 & \cellcolor[HTML]{ABDDC5}62.3 $\pm$ 0.8 & \cellcolor[HTML]{D7EFE4}62.2 $\pm$ 0.7 \\
$P^{lin}$    & \cellcolor[HTML]{DBF1E6}80.5 $\pm$ 0.5 & \cellcolor[HTML]{AFDFC8}80.9 $\pm$ 0.7 & \cellcolor[HTML]{F1FAF6}80.3 $\pm$ 0.7 & \cellcolor[HTML]{D4EEE1}62.2 $\pm$ 0.5 & \cellcolor[HTML]{96D5B6}62.8 $\pm$ 0.2 & \cellcolor[HTML]{B7E2CD}62.3 $\pm$ 0.8 \\
$P^{sig}$    & \cellcolor[HTML]{FFFFFF}80.2 $\pm$ 0.7 & \cellcolor[HTML]{88CFAC}81.5 $\pm$ 0.7 & \cellcolor[HTML]{C9E9DA}80.7 $\pm$ 1.1 & \cellcolor[HTML]{FAFDFC}62.2 $\pm$ 1.6 & \cellcolor[HTML]{8DD1B0}62.9 $\pm$ 0.8 & \cellcolor[HTML]{94D4B5}62.8 $\pm$ 0.4 \\
$P^{lin-ms}$ & \cellcolor[HTML]{CDEBDC}80.6 $\pm$ 0.6 & \cellcolor[HTML]{ABDDC5}80.9 $\pm$ 0.9 & \cellcolor[HTML]{7DCBA5}81.7 $\pm$ 1.3 & \cellcolor[HTML]{BAE3D0}62.3 $\pm$ 0.4 & \cellcolor[HTML]{C6E8D7}62.3 $\pm$ 0.6 & \cellcolor[HTML]{A5DBC1}62.5 $\pm$ 0.9 \\
$P^{sig-ms}$ & \cellcolor[HTML]{DFF2E9}80.5 $\pm$ 0.6 & \cellcolor[HTML]{71C69C}81.9 $\pm$ 0.8 & \cellcolor[HTML]{57BB8A}82.3 $\pm$ 0.9 & \cellcolor[HTML]{7CCAA4}63.2 $\pm$ 0.6 & \cellcolor[HTML]{69C397}63.6 $\pm$ 0.7 & \cellcolor[HTML]{57BB8A}64.0 $\pm$ 0.6 \\
\hline
\multicolumn{7}{c}{VSE$\infty$(WSL)}\\
\hline
$P^{con}$    & \cellcolor[HTML]{FFFFFF}87.9 $\pm$ 1.1 & \cellcolor[HTML]{F1B3AE}89.4 $\pm$ 1.3 & \cellcolor[HTML]{FBE6E4}88.4 $\pm$ 0.6 & \cellcolor[HTML]{FFFFFF}74.0 $\pm$ 0.4 & \cellcolor[HTML]{FBE8E6}74.5 $\pm$ 0.6 & \cellcolor[HTML]{FFFFFF}74.0 $\pm$ 0.2 \\
$P^{lin}$    & \cellcolor[HTML]{FEF7F7}88.0 $\pm$ 0.1 & \cellcolor[HTML]{F7D4D1}88.8 $\pm$ 0.4 & \cellcolor[HTML]{F5CAC6}89.0 $\pm$ 0.6 & \cellcolor[HTML]{FFFCFC}74.1 $\pm$ 0.7 & \cellcolor[HTML]{F8D8D6}74.8 $\pm$ 0.6 & \cellcolor[HTML]{FBE7E5}74.5 $\pm$ 0.7 \\
$P^{sig}$    & \cellcolor[HTML]{FAE1DF}88.5 $\pm$ 0.7 & \cellcolor[HTML]{F7D5D3}88.7 $\pm$ 0.9 & \cellcolor[HTML]{EEA59F}89.7 $\pm$ 0.4 & \cellcolor[HTML]{F8D7D5}74.8 $\pm$ 0.7 & \cellcolor[HTML]{F3BDB9}75.4 $\pm$ 0.2 & \cellcolor[HTML]{F4C4C0}75.2 $\pm$ 0.2 \\
$P^{lin-ms}$ & \cellcolor[HTML]{FEF7F7}88.0 $\pm$ 0.6 & \cellcolor[HTML]{F0AFA9}89.5 $\pm$ 0.7 & \cellcolor[HTML]{EFAAA4}89.6 $\pm$ 0.8 & \cellcolor[HTML]{FAE4E2}74.6 $\pm$ 0.7 & \cellcolor[HTML]{F1B3AE}75.6 $\pm$ 0.7 & \cellcolor[HTML]{F5C9C5}75.1 $\pm$ 0.4 \\
$P^{sig-ms}$ & \cellcolor[HTML]{F1B3AE}89.4 $\pm$ 0.8 & \cellcolor[HTML]{EFABA6}89.6 $\pm$ 0.3 & \cellcolor[HTML]{E67C73}90.6 $\pm$ 0.8 & \cellcolor[HTML]{EEA39C}75.9 $\pm$ 0.6 & \cellcolor[HTML]{ED9E97}76.0 $\pm$ 0.3 & \cellcolor[HTML]{E67C73}76.7 $\pm$ 0.2 \\
\end{tabular}
\end{adjustbox}
\caption{Result of Image$ \rightarrow$ Text and Text $\rightarrow$ Image Recall@1 Flickr 30K test with different gradient combinations on VSE$\infty$(BUTD) and VSE$\infty$(WSL).}
\label{table:vseinfty}
\end{table}